\documentclass{article}

\usepackage{arxiv}

\usepackage[utf8]{inputenc}
\usepackage[T1]{fontenc}
\usepackage{hyperref}
\usepackage{url}        
\usepackage{booktabs}   
\usepackage{amsfonts}   
\usepackage{nicefrac}   
\usepackage{microtype}  
\usepackage{lipsum}		
\usepackage{graphicx}
\usepackage[numbers,sort,compress]{natbib}
\usepackage{doi}
\usepackage{xspace}
\usepackage{amsmath}
\usepackage{xcolor} 
\usepackage[capitalise,noabbrev]{cleveref} 
\usepackage{booktabs}
\usepackage{multirow}
\usepackage{array}
\usepackage[absolute,overlay]{textpos}
\usepackage{caption}
\usepackage{subcaption}
\usepackage{mathtools}
\usepackage{tikz}
\usepackage{wrapfig}

\usepackage{abbr_style}

\hypersetup{
    colorlinks=true,
    linkcolor={red!50!black},
    filecolor=magenta,      
    urlcolor={blue!80!black},
    citecolor={blue!50!black},
    pdftitle={Concept Bottleneck Models Without Predefined Concepts}
    }

\newcommand{\ours}{UCBM\xspace}

\definecolor{cadmiumgreen}{rgb}{0.0, 0.42, 0.24}

\title{Concept Bottleneck Models\\Without Predefined Concepts}

\author{
Simon Schrodi\thanks{Equal contribution.}\hspace{5mm} Julian Schur$^\ast$\hspace{5mm} Max Argus\hspace{5mm} Thomas Brox \vspace{2mm} \\
  University of Freiburg\hspace{4mm}
  \vspace{1mm} \\
  $\texttt{\{schrodi,schurj,argusm,brox\}@cs.uni-freiburg.de}$
}

\date{}

\renewcommand{\shorttitle}{Concept Bottleneck Models Without Predefined Concepts}

\begin{document}
\maketitle

\begin{abstract}
There has been considerable recent interest in interpretable concept-based models such as Concept Bottleneck Models (CBMs), which first predict human-interpretable concepts and then map them to output classes. To reduce reliance on human-annotated concepts, recent works have converted pretrained black-box models into interpretable CBMs post-hoc. However, these approaches predefine a set of concepts, assuming which concepts a black-box model encodes in its representations. In this work, we eliminate this assumption by leveraging unsupervised concept discovery to automatically extract concepts without human annotations or a predefined set of concepts. We further introduce an input-dependent concept selection mechanism that ensures only a small subset of concepts is used across all classes. We show that our approach improves downstream performance and narrows the performance gap to black-box models, while using significantly fewer concepts in the classification. Finally, we demonstrate how large vision-language models can intervene on the final model weights to correct model errors.
\end{abstract}

\section{Introduction}
Deep neural networks have achieved tremendous success in a wide range of tasks and modalities, such as vision and natural language. However, they are \emph{black-box} models, making it difficult for humans to understand and comprehend their decisions.
Thus, there has been a profound recent interest in developing \emph{interpretable} models. One popular framework are Concept Bottleneck Models (CBMs) \citep{koh2020concept}, \ie, models that first predicts human-interpretable concepts and then use these concepts to predict the classes \citep{lampert2009learning,kumar2009attribute}. 
CBMs are trained in an end-to-end fashion through supervision on \emph{both} the concepts and classes.
However, the need for human-annotated concepts during model training requires time-consuming and expensive collection of such annotations.

To address the above limitation of supervised CBMs, recent work \citep{yuksekgonul2023posthoc,oikarinen2023labelfree,marcinkevivcs2024beyond} has converted pretrained, black-box models into CBMs in a \emph{post-hoc} fashion.
These works either learned linear probes on human-annotated concepts or used vision-language text features to align a black-box model's activations with human-interpretable concepts.
Note that the linear probing approach still requires human-annotated concepts, but they are only needed during the conversion stage.
However, a crucial assumption underlies all the aforementioned approaches: it is implicitly assumed that the predefined concepts are encoded by some pretrained, black-box model. 
But
\begin{center}
    \emph{how can we determine a priori which concepts some pretrained, black-box model encodes?}
\end{center}

To eliminate this assumption, we propose to extract concepts in an \emph{unsupervised} manner instead.
Specifically, we draw from the rich literature on unsupervised concept discovery \citep{ghorbani2019towards,zhang2021invertible,zou2023transparency,fel2023craft,vielhaben2023multidimensional,fel2023holistic,huben2024sparse,stein2024towards}. The main idea of these approaches is to find a (small) set of concepts that faithfully represents a model's activations. \citet{fel2023holistic} has shown that many of these approaches can be seen as an instance of the dictionary learning problem \citep{dumitrescu2018dictionary}. That is, these approaches aim to find a dictionary matrix (\ie, a dictionary of concepts) that defines the (concept) basis vectors while the sparse coefficient matrix weighs these basis vectors.
After learning such a dictionary matrix, we propose to project a black-box model's representation onto the space spanned by the concept basis vectors of this dictionary matrix.
It is important to observe that there is no need for human-annotated concepts nor to predefine concepts a priori. Instead, we effectively shift the concept selection step to after the concept discovery.

After constructing the (concept) space spanned by the human-interpretable concepts, we use a sparse linear model to map from that space to the class output space. Following previous work \citep{yuksekgonul2023posthoc,oikarinen2023labelfree}, we apply sparsity to the weights of the linear layer. This results in a sparsity \emph{per-class}, \ie, the decision for a class relies only a small set of concepts.
However, we found that it overlooks the number of concepts used \emph{across all classes}. For example, while class $i$ may not use concept $k$, another class $j$ might use it for a given input sample. Consequently, there are many more concepts (typically $\geq 90\%$ of the available concepts) that influence the entire classification.
If a concept should not be used in classification, either its projection needs to be zero, or the column vector for that concept needs to be the null vector. We typically did not observe the former. Note that the latter is \emph{input-independent} and, thus, effectively reduces the number of concepts and as a result downstream performance (\cref{fig:concepts_vs_performance} shows the relation between both).

To address this challenge, we propose a simple \emph{input-dependent concept selection mechanism} that enforces additional sparsity on the number of concepts used across all classes. Technically, we employ a gating mechanism that selectively removes concepts, retaining only a sparse set of them per input. To enforce sparsity on the number of concepts used, we apply a sparsity regularization to the output of the gating mechanism. It is important to observe that this mechanism is \emph{dynamic}, meaning the concepts that are removed or retained are input-dependent.

Our contributions are as follows:
\begin{itemize}
    \item We propose a novel method (\cref{fig:method_overview}), which we dub Unsupervised Concept Bottleneck Models (UCBMs), as it converts any pretrained model into a concept-based model \emph{without} requiring human-annotated concepts or predefining the set of concepts. This allows \ours to freely find concepts that most faithfully represent the activations of the pretrained black-box model.
    \item Further, we propose a novel input-dependent concept selection mechanism. It retains a sparse set of concepts, thereby substantially increasing the sparsity across all output classes. For example, as few as ca. $0.7\%$ of the available concepts are used per input (see \cref{tab:sparsity}).
    \item Our experiments show that \ours can further close the gap to their black-box counterparts, while having a substantially higher degree of sparsity.
    \item Lastly, we showcase a novel way of utilizing large language-vision models to inform model editing.
\end{itemize}
\section{Unsupervised Concept Bottleneck Models}
\begin{figure}[t]
    \centering
    \includegraphics[width=1\textwidth]{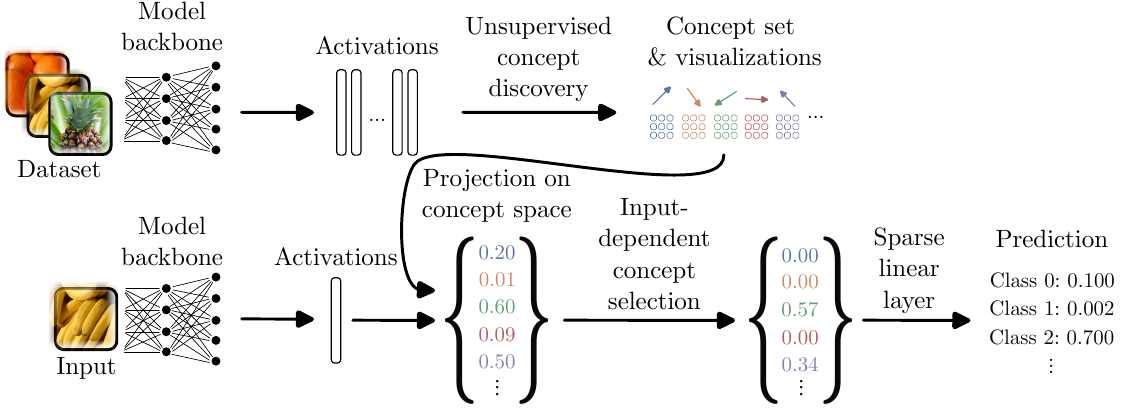}
    \caption{\textbf{Overview of \ours.} In the first step (top), we extract concepts (represented as normalized vectors) using an unsupervised concept discovery method. Given an input sample, we project the activations on the space spanned by these concept vectors (middle at the bottom). Finally, we train an interpretable classifier consisting of an input-dependent concept selection mechanism and sparse linear layer (middle to right at the bottom).}
    \label{fig:method_overview}
\end{figure}
Our approach, Unsupervised Concept Bottleneck Models (UCBMs), consists of two main steps: (1) learning a concept space through unsupervised concept discovery (\cref{sub:concept_subspace}) and (2) learning an interpretable classifier via supervision on top of that concept space (\cref{sub:interpretable_predictor}).
\Cref{fig:method_overview} provides an overview of \ours and both steps are described in detail below.

It is important to note that \ours does \emph{not} require any human-annotated concepts, nor does it require a predefined set of concepts.
This is in contrast to prior post-hoc methods that required predefined concepts either through the set of human-annotated concepts \citep{yuksekgonul2023posthoc,marcinkevivcs2024beyond} or precomputed text features of a vision-language model \citep{yuksekgonul2023posthoc,oikarinen2023labelfree,marcinkevivcs2024beyond}.
Moreover, the choice of training samples with human-annotated concepts impacts the extracted concepts \citep{ramaswamy2023overlooked} or results in entangled concepts \citep{nicolson2024explaining}.
In contrast, \ours does not assume a priori which concepts the pretrained, black-box model encodes but extracts the ones that most faithfully reconstruct its activations \citep{fel2023craft,fel2023holistic}.

Besides the above, we introduce a novel notion of sparsity for concept-based models in the second step of \ours. Previously, sparsity was applied solely to the weights of the linear model. We found that while this led to sparsity per-class, it did not lead to a sparsity across all classes, \cf, \cref{tab:sparsity}. That is, it does not take the total number of concepts across all classes into account. 
As a remedy, we propose an input-dependent concept selection mechanism that enforces that only a sparse set of concepts influence the classification across all classes.

\paragraph{Notations.}
Let $f:\mathcal{X}\rightarrow\mathbb{R}^p$ be a pretrained, black-box model that maps from an input space $\mathcal{X}\subseteq\mathbb{R}^d$ to an embedding space of a size of $p$. Further, let $\mathbf{X}\in\mathbb{R}^{n\times d}$ be the input data matrix where the $i^{th}$ row is the input $x_i\in\mathcal{X}$ and let $\mathbf{A} = f(\mathbf{X}) \in \mathbb{R}^{n \times p}$ be the activations. Lastly, $\mathcal{Y}$ denotes the binary or categorical class label space.

\subsection{Learning the concept space}\label{sub:concept_subspace}
\paragraph{Unsupervised concept discovery.}
The goal of unsupervised concept discovery is to extract a small set of interpretable concepts $c\in C$.
It is popular in concept-based interpretability to assume a linear relationship between the concepts and activations $\mathbf{A}$ \citep{kim2018interpretability,elhage2022toy}, \ie, $\mathbf{C}\in\mathbb{R}^{p\times k}$, where $k$ is the number of concepts.
As highlighted by \citet{fel2023holistic}, many unsupervised concept discovery methods can be seen as (low-rank) approximation of the activations formulated as an instance of the dictionary learning problem \citep{dumitrescu2018dictionary}: 
\begin{equation}\label{eq:dictionary_learning_problem}
    (\mathbf{U}^*, \mathbf{C}^*) = \underset{{\mathbf{U}, \mathbf{C}}}{\arg \min} ~ ||\mathbf{A} - \mathbf{U} \mathbf{C}^T||_F^2 \qquad ,
\end{equation}
where $\mathbf{U}\in\mathbb{R}^{n\times k}$ is a representation of the activations $\mathbf{A}$ \wrt a new basis (sparse coefficient matrix) spanned by the set of concept activation vectors $\mathbf{C}$ (dictionary matrix), and $|| \cdot ||_F$ denotes the Frobenius norm.
Recent unsupervised concept discoverey methods differ in the constraints they apply on $\mathbf{U},\mathbf{C}$ in \cref{eq:dictionary_learning_problem}:
K-Means ($u_i\in\{e_1,...,e_k\}~\forall i$) \citep{ghorbani2019towards},
PCA ($\mathbf{C}^T\mathbf{C}=\mathbf{I}$) \citep{zhang2021invertible,zou2023transparency},
non-negative matrix factorization ($\mathbf{U}\geq 0,\mathbf{C}\geq 0$) \citep{zhang2021invertible,fel2023craft}, or
sparse autoencoders ($\mathbf{U}=\psi(\mathbf{A}),||\mathbf{U}||_0\leq k$) \citep{makhzani2014k,huben2024sparse}.

We chose non-negative matrix factorization for \ours. Note that our approach will benefit from improvements of unsupervised concept discovery methods such as the recently popular sparse autoencoders.

\paragraph{How faithful are the discovered concepts?}
\begin{figure}[t]
    \begin{subfigure}[b]{0.48\linewidth}
        \includegraphics[width=1\linewidth]{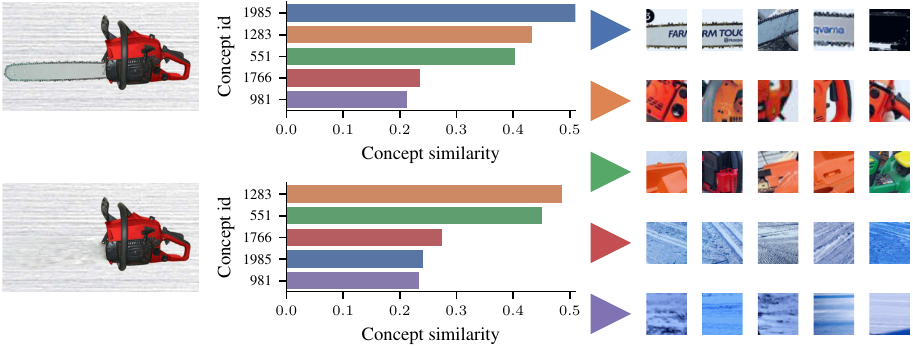}
        \caption{Effect of removing the saw blade (concept id 1985) from a chainsaw.}
        \label{subfig:concept_removal_chainsaw}
    \end{subfigure}
    \hfill
    \begin{subfigure}[b]{0.48\linewidth}
        \includegraphics[width=1\linewidth]{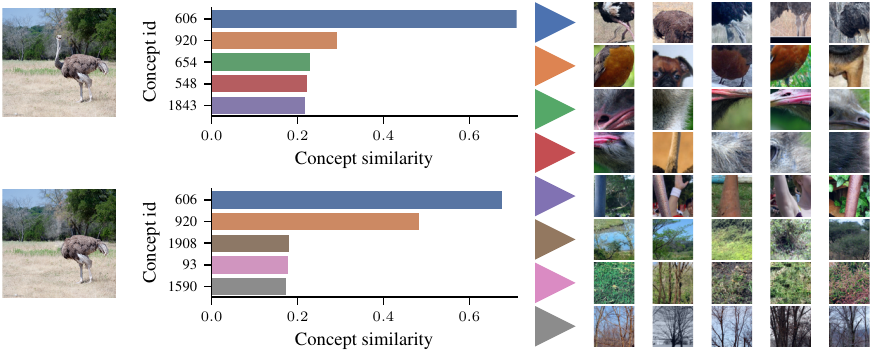}
        \caption{Effect of removing head and neck (concept id 654, concept id 548) of an ostrich.}
        \label{subfig:concept_removal_ostrich}
    \end{subfigure}
    \caption{\textbf{Concepts discovered in an unsupervised manner exhibit faithful behavior.}
    We remove image parts and observe the change in activation-concept cosine similarities for a chainsaw (\subref{subfig:concept_removal_chainsaw}) and an ostrich (\subref{subfig:concept_removal_ostrich}).}
    \label{fig:removed_concept}
\end{figure}
To show that discovered concepts behave in a faithful manner, we analyzed the change in cosine similarities between activations and concepts under removal of relevant image parts for that concept; see \cref{fig:removed_concept}.
In \cref{subfig:concept_removal_chainsaw}, cosine similarity significantly reduces for the saw blade (concept id 1985).
In \cref{subfig:concept_removal_ostrich}, cosine similarity for the ostrich's neck and head (concept id 645 \& 548) even disappear from the top-5 cosine similarity scores. 
These results indicate the faithfulness of the discovered concepts. For a thorough analysis, please refer to, \eg, \citet{fel2023holistic}.

\subsection{Learning the interpretable classifier}\label{sub:interpretable_predictor}
Previous work showed that sparse linear models are more interpretable than dense linear models \citep{wong2021leveraging}. Sparsity is applied to the weights, resulting in the classification of a particular class being based on only a few features (in our case concepts). 
However, we find that the total number of concepts that are used for the classification across all classes is barely affected by the sparsity.
In fact, we find that most concepts are used in the classification; see \cref{tab:sparsity}.

To also enforce sparsity on the total number of concepts across all classes, we propose a simple input-dependent concept selection mechanism that selectively removes concepts and only retains a small set of concepts.
As a result, classification across all classes will use a substantially fewer number of concepts across all classes. Besides that, the input-dependency of the mechanism also accounts for cases where concepts are not visible in the input image, \eg, due to occlusions.
Below, we first revisit sparsity applied to the weights and then describe our mechanism to achieve sparsity across all classes.

\paragraph{Sparse weights in the linear model.}
We followed previous work and employed sparsity to the weights of a linear model \citep{wong2021leveraging,yuksekgonul2023posthoc,oikarinen2023labelfree}:
\begin{equation}\label{eq:class_level_sparsity}
    \min_{\mathbf{W}, b} \sum_{i=1}^N \mathcal{L}(\mathbf{W}\text{proj}_C f(x_i) + b, y_i) + \lambda_w R_{\alpha}(\mathbf{W}) \qquad ,
\end{equation}
where $\text{proj}_{\mathbf{C}} f(x_i)$ denotes the normalized cosine similarity between activations and concepts,
$\mathbf{W}\in\mathbb{R}^{|\mathcal{Y}|\times k}$ denotes the weights,
$b\in\mathbb{R}^{|\mathcal{Y}|}$ denotes the bias, $y_i\in\mathcal{Y}$ denotes the target class for the $i$th sample for a total of $|\mathcal{Y}|$ classes, 
$\mathcal{L}$ denotes the task-specific loss function,
$\lambda_w$ denotes the regularization strength on the weights $\mathbf{W}$, and
$R_{\alpha}(\mathbf{W}):=(1-\alpha)\frac{1}{2}||\mathbf{W}||_F + \alpha ||\mathbf{W}||_{1,1}$ denotes the elastic net regularization \citep{zou2005regularization}.
Note that $\text{proj}_{\mathbf{C}} f(x_i)$ is frozen.
The sparsity leads to classes relying only on a small set of concepts for their decisions. However, we found that it does not lead to few concepts being used across all classes.

\paragraph{Input-dependent concept selection mechanism.}
To enforce that only a small set of concepts is used across all classes, we propose an input-dependent concept selection mechanism consisting of a gate and sparsity regularization applied to the output of that gate.
Specifically, we define the gate as follows:
\begin{equation}\label{eq:gating}
    \pi(x_i):=\max(0,\text{proj}_{\mathbf{C}} f(x_i)-o) \qquad ,
\end{equation}
where $o\in\mathbb{R}_{+}^{k}$ denotes a trainable offset parameter.
To enforce that only few concepts are used per input $x_i$, we would like to minimize $||\pi(x_i)||_0 ~\forall x_i\in\mathcal{X}$. However, since L0 is non-differentiable, we employ the elastic net regularization on the output of the gate $\pi(\cdot)$.
As a result, our proposed mechanism removes concepts that are unlikely to be present in the input (low normalized cosine similarities) and only retains a small set of concepts per input sample.

\paragraph{Final interpretable classifier.}
Combining \cref{eq:gating} with \cref{eq:class_level_sparsity}, we obtain the final interpretable classifier:
\begin{equation}
    \min_{\mathbf{W}, b} \sum_{i=1}^N \mathcal{L}(\mathbf{W}\pi(x_i) + b, y_i) + \lambda_\pi R_{\alpha}(\pi(x_i)) + \lambda_w R_{\alpha}(\mathbf{W}) \qquad ,
\end{equation}
where $\lambda_\pi$ governs the regularization strength of the sparsity on the number of concepts used in the entire classification decision per input sample.
It is important to observe that the classification decision is still linear \wrt the \emph{active} concepts $\pi(x_i)>0$ for input $x_i$.

\paragraph{Concept dropout.}
Finally, we found that the sparsity on the weights leads the interpretable classifier to over-rely on a single concept.
To reduce over-reliance, we added a dropout layer \citep{srivastava2014dropout} on the output of the gate $\pi(\cdot)$. Note that dropout is applied per concept. Consequently, the interpretable classifier is encouraged to spread its classification across more concepts.
Interestingly, we found that this can even improve downstream performance in our experiments.
\section{Experiments}
We evaluated \ours on diverse image classification datasets and compared it to relevant baselines. We show that \ours achieves better downstream performance to prior works and narrows the gap to their black-box counterparts, while relying on substantially less concepts in its classification. Lastly, we showcase how recent large vision-language models can be leveraged to inform weight editing of the interpretable classifier to fix model errors.

\paragraph{Datasets \& backbone models.}
We evaluated our approach on ImageNet \citep{deng2009imagenet} with ResNet-50 V2 \citep{he2016deep}, CUB \citep{wah2011caltech} with ResNet-18 pretrained on CUB\footnote{Provided by \url{https://github.com/osmr/imgclsmob}.}, and Places-365 \citep{zhou2017places} with ResNet-18 pretrained on Places-365\footnote{Provided by \citet{oikarinen2023labelfree} at \url{https://github.com/Trustworthy-ML-Lab/Label-free-CBM}.}.
These datasets cover a diverse set of tasks from standard image classification (ImageNet), fine-grained classification (CUB), to scene recognition (Places-365). The datasets also have very different amounts of training samples (from ca. 5000 to 2~million).

\paragraph{Experimental setup.}
Following \citet{oikarinen2023labelfree}, we set the number of concepts proportional to the number of classes for each task.
Since we have found that the number of concepts substantially influence downstream performance, we used various proportionality factors $\{0.5,1,3,5\}$.
All models were trained on a single NVIDIA RTX 2080 GPU and a full training run takes from few minutes to a maximum of 1-2 days depending on dataset size and number of concepts.
We report top-1 accuracy throughout our experiments.

\paragraph{Implementation details.}
We trained our method with Adam using an initial learning rate of $0.001$ for ImageNet and Places-365, $0.01$ for CUB, and cosine annealing learning rate scheduling for all of them. We used $\lambda_{\pi} = 1e-4$ for CUB and $\lambda_{\pi} = 2e-5$ for ImageNet and Places-365. Further, we penalized the final layer's weights with $\lambda_w = 1e-4$ for ImageNet, $8e-4$ for CUB and $4e-4$ for Places-365. We set $\alpha = 0.99$ for the elastic net regularization and the dropout rate to $0.2$.
The reason for choosing these regularization strengths was that they yielded a good trade-off between downstream performance and sparsity. Our sensitivity analysis in \cref{fig:sensitivity_analysis} provides a comprehensive understanding of the effect of each hyperparameter on the trade-off between downstream performance and sparsity.

\paragraph{Baselines.}
We compared \ours to Post-hoc CBM \citep{yuksekgonul2023posthoc} and Label-free CBM \citep{oikarinen2023labelfree}, as these are the most related to our work. Note that Post-hoc CBM requires annotated concept data to learn the concept space, while Label-free CBM leverage CLIP's text features. Importantly, both approaches require to predefine a concept set a priori.

\subsection{Downstream Performance and Sparsity Results}
\begin{figure}[t]
    \centering
    \includegraphics{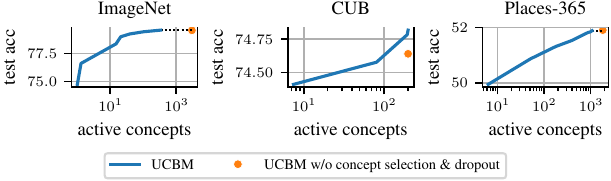}
    \caption{\textbf{Worse performance of \ours to \ours without concept selection \& dropout in \cref{tab:downstream_performance} is due to the increased sparsity.} We plot the mean number of active concepts per input as we make $\lambda_\pi$ smaller and reduce the dropout rate (we only plot the Pareto-optimal points). The dotted black lines serve only as visual guides.
    We find that \ours closes the gap to \ours without concept selection \& dropout as we move towards less sparsity.}
    \label{fig:performance_vs_active_concepts}
\end{figure}

\paragraph{How does \ours compare to the baselines?}
\begin{table}[t]
    \centering
    \caption{\textbf{\ours achieves superior performance to the baselines and narrows the gap to the original black-box model.} We report mean top-1 accuracy with standard error over three training runs (we kept the learned concept space fixed throughout). 
    Best numbers are shown in \textbf{bold} and second best in \textit{italic}.
    We used 3000 (ImageNet), 200 (CUB), and 1825 (Places-365) concepts for both \ours variants.
    Note that the baselines use a different number of concepts: Post-hoc CBM (CUB: 112), Label-free CBM after its concept removal steps (ImageNet: 4521, CUB: 212, Places-365: 2008). This makes direct comparison hardly possible. 
    Thus, \cref{fig:concepts_vs_performance} takes the number of concepts into account and shows that \ours \emph{Pareto-dominates} the baselines. Note that the concept selection mechanism trades-off downstream performance for substantially higher sparsity (\cref{tab:sparsity}). This explains the difference of \ours to \ours w/o concept selection \& concept dropout (\cref{fig:performance_vs_active_concepts}).}
    \label{tab:downstream_performance}
    \begin{tabular}{lcccc}
    \toprule
         &  \multicolumn{3}{c}{Dataset} \\
         \cline{2-4}\\
        Approach & ImageNet & CUB & Places-365 \\\midrule
        Original black-box model & 80.9 & 76.7 & 53.69\\\midrule
        Post-hoc CBM \citep{yuksekgonul2023posthoc} & n/a & 58.8 & n/a\\
        Label-free CBM \citep{oikarinen2023labelfree} & 78.09 & 74.38  & 50.67 \\\midrule
        \ours & \textit{79.03 $\pm$ 0.03} & \textit{74.58 $\pm$ 0.03} & \textit{50.86 $\pm$ 0.02}  \\
        \ours w/o concept selection \& concept dropout & \textbf{79.60 $\pm$ 0.02} & \textbf{74.64 $\pm$ 0.05} & \textbf{51.90 $\pm$ 0.01} \\\bottomrule
    \end{tabular}
\end{table}
\begin{figure}[t]
    \centering
    \includegraphics[width=0.75\linewidth]{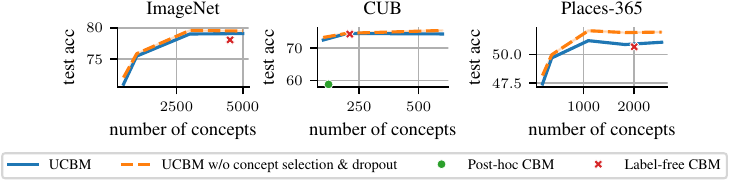}
    \caption{\textbf{\ours Pareto-dominates the baselines.} We varied the number of available concepts. As expected, we found that the more available concepts the better the downstream performance. Importantly, \ours Pareto-dominates the baseline methods.}
    \label{fig:concepts_vs_performance}
\end{figure}
\Cref{tab:downstream_performance} compares the performance of \ours to the baseline methods across all datasets. 
We find that \ours achieves superior performance to the baselines and narrows the gap to the original black-box model.
Besides that, we find that \ours without concept selection \& dropout achieves better downstream performance than \ours (with concept selection). However, this is to be expected, as \ours uses substantially fewer concepts in the classification (\cref{tab:sparsity}) and \ours indeed closes the gap to \ours without concept selection \& dropout as we allow for less sparsity (\cref{fig:performance_vs_active_concepts}).

During experimentation, we found that downstream performance is influenced by the number of concepts that are used to construct the concept space. Thus, we varied the number of concepts to assess its impact on downstream performance; see \cref{fig:concepts_vs_performance}.
As expected, we find that an increasing number of concepts yields improved downstream performance.
Besides that, we observe that \ours Pareto-dominates the baselines, indicating that its superiority from \cref{tab:downstream_performance} does not stem from the chosen number of concepts.

\paragraph{How sparse are the models' decisions?}\label{par:sparsity}
To understand the sparsity of the models, we computed the mean number of concepts that are used for classification per input sample.
To do this, we computed the mean number of concepts that are retained from the concept selection and also included concepts with a cosine similarity of zero (\ie, $\text{proj}_{\mathbf{C}_j} f(x_i)\approx 0$) as well as ones that are never used in the classification (\ie, null column vectors in the weights of the linear layer).\footnote{We used a small threshold of $1e-5$ to assess whether cosine similarity or the weights are approximately zero, following \citet{oikarinen2023labelfree}.}

\cref{tab:sparsity} shows that the concept selection mechanism leads to substantially less concepts that are used for classification per input sample. For instance on ImageNet, \ours only uses 21 concepts on average (0.7\% of the total number of available concepts). In contrast, Label-free CBM and \ours without concept selection (\& dropout) use on average 4238.0 (93.74\%) or 2999.5 concepts (99.98\%), respectively.
Similar stark differences in sparsity can be observed on CUB and Places-365.
Thus, we posit that this explains the downstream performance advantage of \ours without concept selection \& dropout over \ours (with concept selection) in \cref{tab:downstream_performance}.
To confirm this, we reduced the sparsity by reducing $\lambda_\pi$ as well as the dropout rate. Indeed, we find that the gap closes; see \cref{fig:performance_vs_active_concepts}.
\begin{table}[t]
    \centering
    \caption{\textbf{Our proposed input-dependent concept selection mechanism leads to substantially less concepts being used for the classification.} We report the mean number of concepts (with standard deviation) that are used for classification. The available concepts are shown in parentheses.
    Label-free CBM and \ours without concept selection \& dropout uses almost always all concepts in its classification, while \ours only uses a small amount of the concepts.}
    \label{tab:sparsity}
    \begin{tabular}{lcccc}
    \toprule
         &  \multicolumn{3}{c}{Dataset} \\
         \cline{2-4}\\
        Approach & ImageNet & CUB & Places-365 \\\midrule
        Label-free CBM \citep{oikarinen2023labelfree} & 4238.0 $\pm$ 0.19 (4521) & 211.9 $\pm$ 0.05 (212) & 1820.0 $\pm$ 0.12 (2008)\\\midrule
        \ours & \textbf{21.0 $\pm$ 0.93} (3000) & \textbf{80.8 $\pm$ 2.56} (200) & \textbf {51.4 $\pm$ 4.78} (1825) \\
        \ours w/o concept selection \& dropout & 2999.5 $\pm$ 5.72 (3000) & 200.0 $\pm$ 0.0 (200) & 1825.0 $\pm$ 0.0 (1825) \\\bottomrule
    \end{tabular}
\end{table}

\begin{figure}[t]
    \centering
    \begin{subfigure}[c]{0.3\linewidth}
        \includegraphics[width=\linewidth]{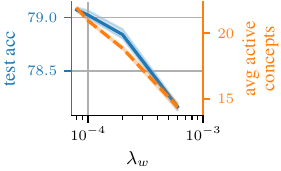}
        \caption{Performance \vs $\lambda_w$.}
        \label{subfig:lam_w_imagenet}
    \end{subfigure}
    \begin{subfigure}[c]{0.3\linewidth}
        \includegraphics[clip,trim=0 0 0 0,width=\linewidth]{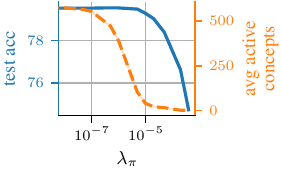}
        \caption{Performance \vs $\lambda_\pi$.}
        \label{subfig:lam_pi_imagenet}
    \end{subfigure}
    \begin{subfigure}[c]{0.3\linewidth}
        \includegraphics[width=\linewidth]{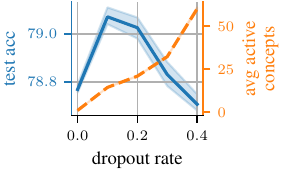}
        \caption{Performance \vs dropout.}
        \label{subfig:dropout_imagenet}
    \end{subfigure}
    \caption{\textbf{Sensitivity analysis over $\lambda_w$ (\subref{subfig:lam_w_imagenet}), $\lambda_\pi$ (\subref{subfig:lam_pi_imagenet}), and the dropout rate (\subref{subfig:dropout_imagenet}) for ImageNet.} Stronger regularization strengths ($\lambda_w$,$\lambda_\pi$) lead to worse downstream task performance. For dropout there exists a sweet spot. Results for the other datasets are provided in \cref{sec:sensitivity_analysis_other}.}
    \label{fig:sensitivity_analysis}
\end{figure}

\paragraph{Sensitivity analysis.}
To understand the impact of the hyperparameters beyond the number of concepts (\cref{fig:concepts_vs_performance}), we ran sensitivity analyses over $\lambda_w$ (\cref{subfig:lam_w_imagenet}), $\lambda_\pi$ (\cref{subfig:lam_pi_imagenet}), and the dropout rate (\cref{subfig:dropout_imagenet}).
\cref{subfig:lam_w_imagenet,subfig:lam_pi_imagenet} show larger regularization lengths lead to worse downstream performance but increase sparsity; except for $\lambda_w$ on Places-365 surprisingly.
\cref{subfig:dropout_imagenet} shows that sparsity reduces as we increase the dropout rate (except for Places-365 again) while downstream performance has some sweet spot.

\subsection{Explainable Individual Decisions}
\begin{figure}[t]
    \centering
    \begin{subfigure}[b]{0.48\linewidth}
        \includegraphics[width=1\linewidth]{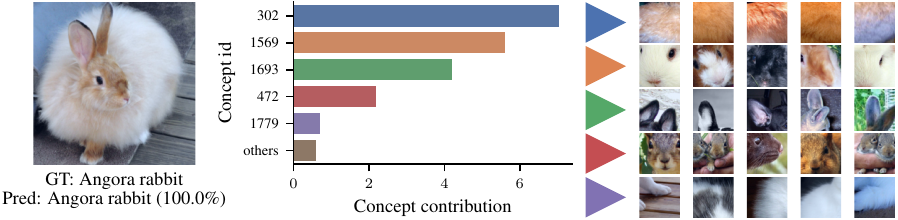}
        \caption{Explainable decision for ImageNet class `Angora rabbit'.}
        \label{subfig:explainable_decision_rabbit}
    \end{subfigure}
    \hfill
    \begin{subfigure}[b]{0.48\linewidth}
        \includegraphics[width=1\linewidth]{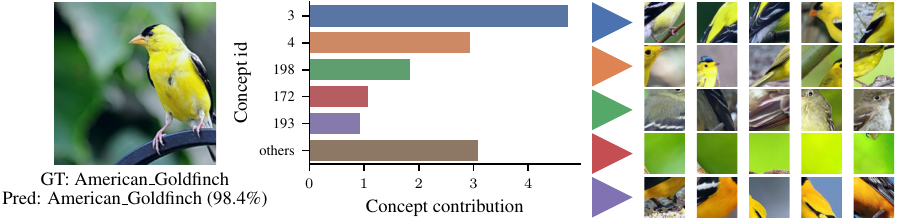}
        \caption{Explainable decision of CUB class `American goldfinch'}
        \label{subfig:explainable_decision_goldfinch}
    \end{subfigure}
    \caption{\textbf{Visualization of two correct decision made by \ours on ImageNet (\subref{subfig:explainable_decision_rabbit}) and CUB (\subref{subfig:explainable_decision_goldfinch}).} More examples are provided in \cref{sec:explainable_decisions_other}.}
    \label{fig:explainable_decisions}
\end{figure}
Concept-based models allow us to understand the classification decision for individual input samples. For this we make use of the fact that the decision is linear \wrt to the active concepts ($\pi(x_i)>0$).
Specifically, we computed the contribution of a concept $j$ for an input sample $i$ to output class $k$ as $\mathbf{W}_{j,k}\text{proj}_{\mathbf{C}_j} f(x_i)$, following \citet{oikarinen2023labelfree}. Note that due to the sparsity (\cf, \cref{tab:sparsity}) most contributions will be zero. We show the top-k concepts with the highest absolute contributions to the classification decision of the predicted class.

\cref{fig:explainable_decisions} and \cref{fig:explainable_decisions_other} in \cref{sec:explainable_decisions_other} illustrate which concepts influence the classification the most.
For instance, \cref{subfig:explainable_decision_rabbit} shows that \ours focuses on rabbit concepts such as fur, face, ears, or pawns.
Similarly, \cref{subfig:explainable_decision_goldfinch} highlights that \ours relies on typical concepts of an American goldfinch such as its bright yellow plumage.

\subsection{Case Study: Fixing Model Errors Through the Feedback of a Large Vision-Language Models}
Finally, we show how we can leverage feedback from a large vision-language model to fix model errors. 
We ran our case study on the first ten classes of ImageNet and labeled all concepts through manual inspection.\footnote{The reason for this is to reduce the costs for concept labeling; see \cref{sec:limitations} for a discussion.}
We fed the large vision-language model (we used GPT-4o \citep{achiam2023gpt}\footnote{\url{https://openai.com/index/hello-gpt-4o/}}) with the misclassified input sample, the top-k corresponding concepts, and their respective model weights with a task description.
Specifically, we tasked the large vision-language model to propose a change of the model weights that flips the incorrect classification to the correct one without impacting other input samples.
We provide an example in \cref{sec:case_study}.
During experimentation, we found that the proposed changes by the large vision-language model were often too strong, resulting in misclassifications of previously correct samples. Thus, we ran a line search with $\beta \in [0, 1]$. Note that we ran line search on the test set to get an upper bound, \ie, the best possible edit. Similar results were achieved when running line search on the training set, while the incorrect instance was always corrected it sometimes led to errors on other instances.

\begin{figure}[t]
    \centering
    \begin{subfigure}[b]{0.48\linewidth}
        \includegraphics[width=1\linewidth]{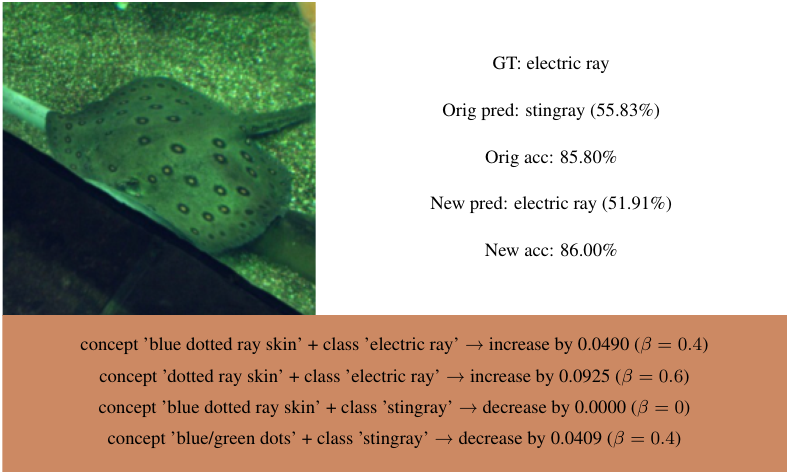}
    \end{subfigure}
    \hfill
    \begin{subfigure}[b]{0.48\linewidth}
        \includegraphics[width=1\linewidth]{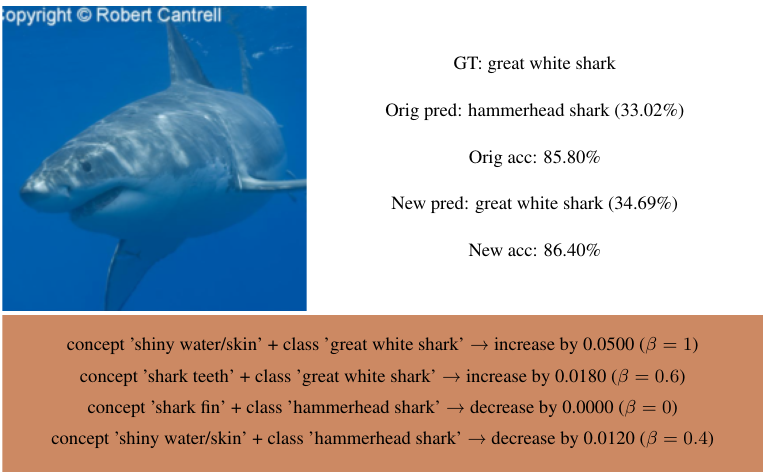}
    \end{subfigure}
    \caption{\textbf{Correcting model weights of \ours using guidance from a large vision-language model.}}
    \label{fig:correcting_weights}
\end{figure}

\cref{fig:correcting_weights} shows two examples that are classified correctly after following the suggestion of the weight edit of the large vision-language model. 
This demonstrates the intervenability of our method. This case study also indicates the potential of large vision-language models to automatically identify and fix the cause of misclassifications. 
\section{Related Works}
\paragraph{Concept-based models.}
The main idea of concept bottleneck models (CBMs) \citep{koh2020concept} is to leverage human-interpretable concepts in their classifications \citep{lampert2009learning,kumar2009attribute}.
Many works highlighted (and partially addressed) the limitations of these types of models \citep{margeloiu2021concept,mahinpei2021promises,havasi2022addressing,marconato2022glancenets,raman2024concept}.
Other works improved the performance-interpretability trade-off \citep{espinosa2022concept,yang2023language} or extended them to image generative models \citep{ismail2023concept} or tabular data \citep{zarlenga2023tabcbm}.
Orthogonal work transformed CLIP \citep{radford2021learning} to a zero-shot, concept-based classifier \citep{menon2023visual}.
Most related to our work are methods that convert a pretrained black-box model into a CBM post-hoc \citep{yuksekgonul2023posthoc,oikarinen2023labelfree,marcinkevivcs2024beyond}. They project the activations of the black-box models' backbone onto a new concept space learned either via supervision or using a vision-language model's text features.
In contrast to prior work, we do not require human-annotated concepts nor a predefined set of concepts.

\paragraph{Concept discovery.}
Early work searched for neuron-aligned concepts \citep{bau2017network,olah2017feature}, while later works, inspired by the superposition hypothesis \citep{elhage2022toy}, went beyond this to (linear) vector \citep{kim2018interpretability,zhou2018interpretable,ghorbani2019towards,zhang2021invertible,zou2023transparency,fel2023craft,huben2024sparse,stein2024towards} or linear subspace \citep{vielhaben2023multidimensional} concept representations.
Early work needed annotated datasets to find concepts that required substantial human effort. Most later works overcame this bottleneck by formulating concept discovery as a dictionary learning problem \citep{fel2023holistic}.
Concurrently, \citet{stein2024towards} compared the representative prowess of discovered concepts of unsupervised concept discovery methods through linear probing.

\paragraph{Model editing.}
The main goal of model editing is to modify a model to remove a bias or rectify model errors.
For instance, it is popular to edit factual knowledge in large language models \citep{zhu2020modifying,meng2022locating}, generative image models \citep{bau2020rewriting,gandikota2023erasing}, or modifying a classifier's prediction rules \citep{santurkar2021editing,oikarinen2023labelfree}.
While previous work relied on, \eg, human intervention or hypernetworks, we leverage recent large vision-language models to inform model editing.
\section{Limitations \& Future Work}\label{sec:limitations}
The main limitation of our approach is the need to label the concepts through manual inspection. Thus, we plotted the most activation images for each concept in our figures.
However, this slows down the step of interpretability by a human, as they need to interpret the most activating images first. While we can manually label the concepts, it does not scale to a large amount of concepts.
Promising approaches for automatic labeling could be an adaption of the proposal-ranker framework \citep{zhong2022describing,dunlap2024describing}, multi-modal large vision-language models \citep{achiam2023gpt}, or a combination of the aforementioned.

Another limitation of our approach is that we only extract concepts from the bottleneck layer of the black-box model.
By extracting concepts across the layers of the black-box model, we can effectively leverage a hierarchy of concepts. For instance, an early layer could find concepts for windows, car body, or wheels, while a later layer could encode a car concept \citep{olah2020zoom}. Since such hierarchy is learned and used by black-box models \citep{zeiler2014visualizing}, it is likely that it will also benefit interpretable concept-based models.

\section{Conclusion}
We presented \ours that converts any pretrained black-box model into an interpretable concept-based model without requiring any human-annotated concepts nor a predefined concept set.
We introduced an input-dependent concept selection mechanism that ensures that only a small set of concepts is used per-class as well as across all classes.
We showed that \ours is superior to the baselines, narrows the gap to black-box models, while being substantially more sparse. Finally, we showcased how we can leverage large vision-language models to inform model editing.

\subsubsection*{Acknowledgments}
\small
This research was funded by the German Research Foundation (DFG) under grant numbers 417962828 and 499552394.
\normalsize

\bibliographystyle{unsrtnat}
\bibliography{main}

\newpage
\appendix
\section{Sensitivity Analysis Results for CUB and Places-365}\label{sec:sensitivity_analysis_other}
\begin{figure}[hb]
    \centering
    \begin{subfigure}[c]{\linewidth}
    \centering
        \includegraphics[width=0.3\linewidth]{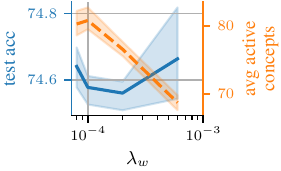}
        \includegraphics[width=0.3\linewidth]{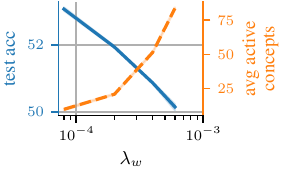}
        \caption{Performance \vs $\lambda_w$.}
        \label{subfig:lam_w_other}
    \end{subfigure}
    \begin{subfigure}[c]{\linewidth}
    \centering
        \includegraphics[clip,trim=0 0 0 0,width=0.3\linewidth]{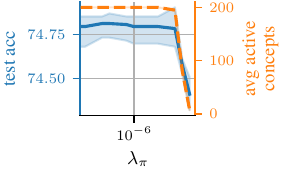}
        \includegraphics[clip,trim=0 0 0 0,width=0.3\linewidth]{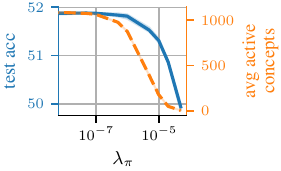}
        \caption{Performance \vs $\lambda_\pi$.}
        \label{subfig:lam_pi_other}
    \end{subfigure}
    \begin{subfigure}[c]{\linewidth}
    \centering
        \includegraphics[width=0.3\linewidth]{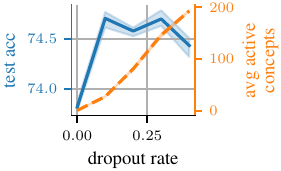}
        \includegraphics[width=0.3\linewidth]{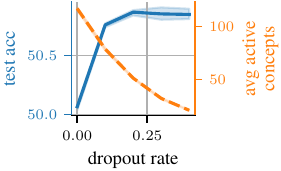}
        \caption{Performance \vs dropout.}
        \label{subfig:dropout_other}
    \end{subfigure}
    \caption{\textbf{Sensitivity analysis over $\lambda_w$ (\subref{subfig:lam_w_other}), $\lambda_\pi$ (\subref{subfig:lam_pi_other}), and the dropout rate (\subref{subfig:dropout_other}) for CUB (left) and Places-365 (right).}}
    \label{fig:sensitivity_analysis_other}
\end{figure}

\cref{fig:sensitivity_analysis_other} provides the results for the sensitivity analysis on CUB and Places-365. Surprisingly, we find sparsity decreases as we increase $\lambda_w$ or the dropout rate for Places-365, while $\lambda_\pi$ behaves as expected.

\section{More Examples of Explainable Decisions}\label{sec:explainable_decisions_other}
\begin{figure}[ht!]
    \centering
    \begin{subfigure}[c]{1\linewidth}
        \includegraphics[width=1\linewidth]{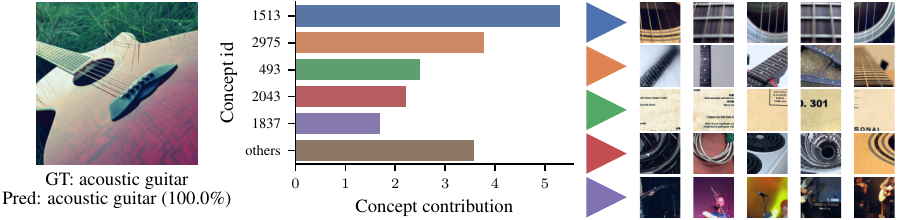}
        \caption{Explainable decision of ImageNet class `acoustic guitar'}
        \label{subfig:explainable_decision_guitar}
    \end{subfigure}
    \begin{subfigure}[b]{1\linewidth}
        \includegraphics[width=1\linewidth]{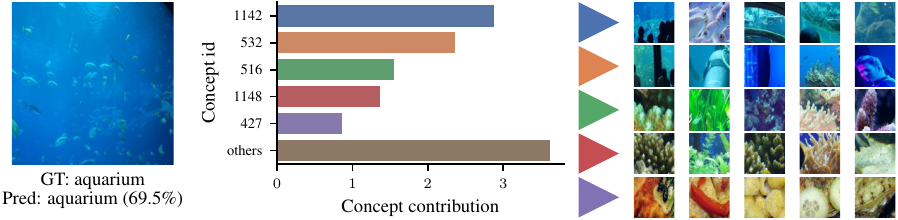}
        \caption{Explainable decision for Places-365 class `aquarium'.}
        \label{subfig:explainable_decision_aquarium}
    \end{subfigure}
    \begin{subfigure}[c]{1\linewidth}
        \includegraphics[width=1\linewidth]{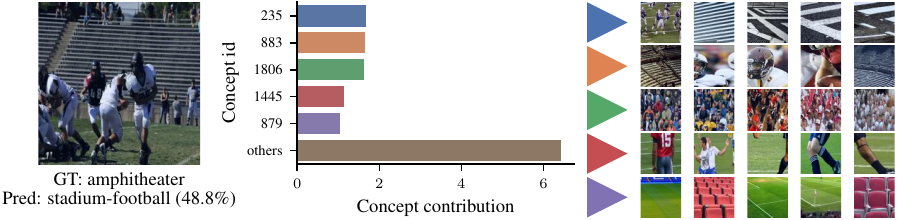}
        \caption{Explainable decision of Places-365 class `amphitheater'}
        \label{subfig:explainable_decision_amphitheater}
    \end{subfigure}
    \begin{subfigure}[c]{1\linewidth}
        \includegraphics[width=1\linewidth]{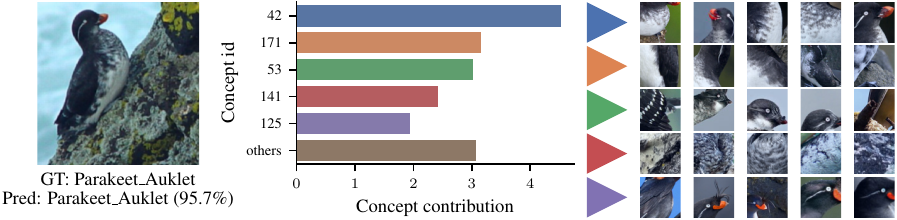}
        \caption{Explainable decision of CUB class `parakeet auklet'}
        \label{subfig:explainable_decision_auklet}
    \end{subfigure}
    \caption{Explainable decisions of ImageNet (\subref{subfig:explainable_decision_guitar}), Places-365 (\subref{subfig:explainable_decision_aquarium},\subref{subfig:explainable_decision_amphitheater}) and CUB (\subref{subfig:explainable_decision_auklet}) classes. }
    \label{fig:explainable_decisions_other}
\end{figure}

\Cref{fig:explainable_decisions_other} provides more examples of explainable decision of \ours. The examples show that \ours uses a small set of human-interpretable concepts for its classification.

\section{Example Input to the Large Vision-Language Model}\label{sec:case_study}
\begin{figure}
\begin{tikzpicture}
\definecolor{chatcolor1}{HTML}{5fedb7}
\definecolor{chatcolor2}{HTML}{b6b8b7}
\fontfamily{cmss}\selectfont
\node[align=left, text width=\linewidth, fill=chatcolor2, rounded corners=1mm, anchor=north west] at (0,0) {
\begin{wrapfigure}{r}{0.4\textwidth}
\includegraphics[width=0.35\textwidth]{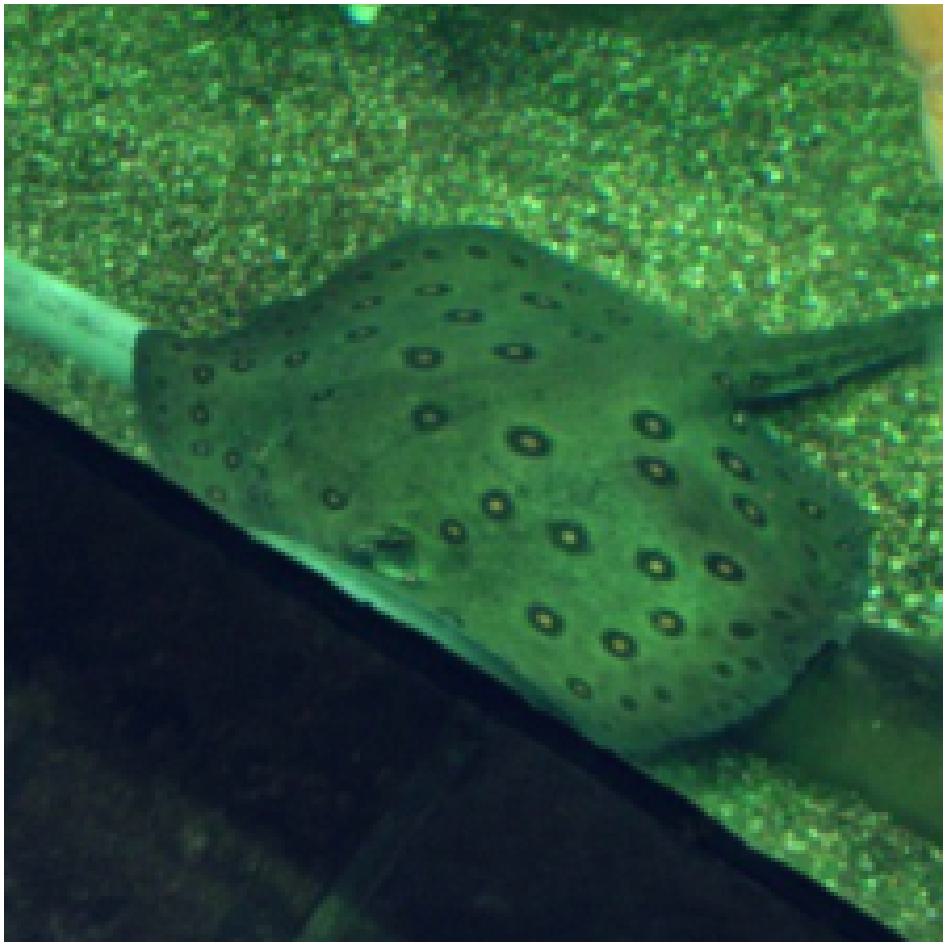}
\centering
\end{wrapfigure}
The image from above (first image) gives the following output if we are putting them into a trained concept bottleneck model. \\
~\\
Ground truth: electric ray\\
Prediction: stingray\\
So the model predicted class stingray, but it should be class electric ray\\
~\\
Biggest five final model outputs sorted by size: \\
Class stingray: 1.97\\
Class electric ray: 1.65\\
Class hammerhead shark: -0.94\\
Class goldfish: -2.81\\
Class tench: -4.27\\
~\\
Largest five concept contributions (by absolute value) for ground truth class electric ray sorted by size (and their weight in the final linear layer): \\
Concept 15: 0.8914 (0.36)\\
Concept 39: 0.6210 (0.48)\\
Concept 44: 0.5858 (0.10)\\
Concept 40: 0.2293 (0.35)\\
Concept 30: -0.1811 (-0.28)\\
~\\
Largest five concept contributions (by absolute value) for predicted class stingray sorted by size (and their weight in the final linear layer): \\
Concept 44: 1.7669 (0.30)\\
Concept 15: 1.6708 (0.68)\\
Concept 39: -0.9347 (-0.72)\\
Concept 26: -0.3604 (-0.21)\\
Concept 22: -0.2038 (-0.17)\\
~\\
The concept contribution therefore is the concept similarity times the weight from the concept and the specific class. \\
The pure concept similarities are given by: \\
Concept 15: 2.4511497020721436\\
Concept 39: 1.2969582080841064\\
Concept 44: 5.871964454650879\\
Concept 40: 0.6597864031791687\\
Concept 30: 0.6427087783813477\\
Concept 26: 1.7485312223434448\\
Concept 22: 1.194166898727417\\
~\\
The concepts are representing the following: \\
Concept 15: blue dotted ray skin\\
Concept 39: dotted ray skin\\
Concept 44: blue/green dots\\
Concept 40: sand ground\\
Concept 30: meadow\\
Concept 26: green\\
Concept 22: ?\\
~\\
Furthermore, the layer from the concept similarities to the prediction is a linear layer. Which weights (concept and class) should I adjust by how much in order to get a correct classification for this image? But be aware to not to change any other classifications. Please give me an answer in case of 'increase/decrease weight of class i and concept i by x'. };
\end{tikzpicture}
\caption{Textual input for correcting linear classifier weights. }
\label{fig:case_study_input}
\end{figure}

\Cref{fig:case_study_input} provides the input to the large vision-language model for the misclassification from the left subfigure in \cref{fig:correcting_weights}.
\cref{fig:case_study_output} provides the answer we have got from the large vision-language model.

\begin{figure}
\begin{tikzpicture}
\definecolor{chatcolor1}{HTML}{5fedb7}
\definecolor{chatcolor2}{HTML}{b6b8b7}
\fontfamily{cmss}\selectfont
\node[align=left, text width=\linewidth,fill=chatcolor1, rounded corners=1mm, anchor=north west] at (120.5,-14.7) {To adjust the weights so that the model correctly classifies the image as an electric ray without affecting other classifications, we'll focus on modifying the weights between the key concepts and the classes. Here's how you can adjust the weights:\\
~\\
    $[\dots]$\\
~\\
\paragraph{Adjustment Calculation:}
~\\~\\
We need to increase the score for the electric ray class while ensuring that the stingray score decreases or does not change much.\\
~\\
\textit{Adjusting Electric Ray Score:}\\
~\\
To increase the electric ray score: $\text{New Weight} = \text{Current Weight} + \Delta$
~\\
\begin{enumerate}
    \item Concept 15:
    \begin{itemize}
        \item Current Contribution: $2.4511 \times 0.36 = 0.8824$
        \item Target Contribution (desired increase): Let's aim for an increase of $0.3$.
        \item $\Delta$ for Concept 15 (Electric Ray): $\Delta_{15} = \frac{0.3}{2.4511} \approx 0.1224$
        \item New Weight for Concept 15 (Electric Ray): $0.36 + 0.1224 = 0.4824$
    \end{itemize}
    \item Concept 39:
    \begin{itemize}
        \item Current Contribution: $1.2970 \times 0.48 = 0.6226$
        \item Target Contribution (desired increase): Let's aim for an increase of $0.2$.
        \item $\Delta$ for Concept 39 (Electric Ray): $\Delta_{39} = \frac{0.2}{1.2970} \approx 0.1542$
        \item New Weight for Concept 39 (Electric Ray): $0.48 + 0.1542 = 0.6342$
    \end{itemize}
\end{enumerate}
~\\
\textit{Adjusting Stingray Score:}\\
~\\
To decrease the stingray score: $\text{New Weight} = \text{Current Weight} - \Delta$
~\\
\begin{enumerate}
    \item Concept 15:
    \begin{itemize}
        \item Current Contribution: $2.4511 \times 0.68 = 1.6667$
        \item Target Contribution (desired decrease): Let's aim for a decrease of $0.5$.
        \item $\Delta$ for Concept 15 (Stingray): $\Delta_{15} = \frac{0.5}{2.4511} \approx 0.2040$
        \item New Weight for Concept 15 (Stingray): $0.68 - 0.2040 = 0.4760$
    \end{itemize}
    \item Concept 44:
    \begin{itemize}
        \item Current Contribution: $5.8720 \times 0.30 = 1.7616$
        \item Target Contribution (desired decrease): Let's aim for a decrease of $0.6$.
        \item $\Delta$ for Concept 44 (Stingray): $\Delta_{44} = \frac{0.6}{5.8720} \approx 0.1022$
        \item New Weight for Concept 44 (Stingray): $0.30 - 0.1022 = 0.1978$
    \end{itemize}
\end{enumerate}
~\\
\paragraph{Summary of Adjustments:}~\\
~\\
\begin{itemize}
    \item Increase weight of Concept 15 (Electric Ray) by $0.1224$.
    \item Increase weight of Concept 39 (Electric Ray) by $0.1542$.
    \item Decrease weight of Concept 15 (Stingray) by $0.2040$.
    \item Decrease weight of Concept 44 (Stingray) by $0.1022$.
\end{itemize}
~\\
These adjustments should help shift the classification from stingray to electric ray for the given image without significantly affecting other classifications.};
\end{tikzpicture}
\caption{Textual output for correcting linear classifier weights. }
\label{fig:case_study_output}
\end{figure}

\end{document}